\documentclass{article}

\usepackage{amssymb, amsmath}

\usepackage[english]{babel}

\usepackage{biblatex}
\usepackage{bbding}
\usepackage{pifont}
\usepackage{wasysym}
\usepackage{authblk}
\usepackage[letterpaper,top=2cm,bottom=2cm,left=3cm,right=3cm,marginparwidth=1.75cm]{geometry}

\usepackage{graphicx}
\usepackage[colorlinks=true, allcolors=blue]{hyperref}
\usepackage{array}
\title{Facial Action Unit Recognition With Multi-models Ensembling}

\author{Wenqiang Jiang \thanks{These authors contributed equally to this work and should be considered co-first authors }}
\author{Yannan Wu \thanks{These authors contributed equally to this work and should be considered co-first authors}}
\author{Fengsheng Qiao \thanks{These authors contributed equally to this work and should be considered co-first authors}}
\author{Liyu Meng}
\author{Yuanyuan Deng}
\author{Chuanhe Liu}
\affil{Beijing Seek Truth Data Technology Co.,Ltd.}
\date{}

\begin{document}
\maketitle
\section{Abstract}

{The Affective Behavior Analysis in-the-wild (ABAW) 2022 Competition gives Affective Computing a large promotion. In this paper, we present our method of AU challenge in this Competition. We use improved IResnet100 as backbone. Then we train AU dataset in Aff-Wild2 on three pertained models pretrained by our private au and expression dataset, and Glint360K respectively. Finally, we ensemble the results of our models. We achieved F1 score (macro) 0.731 on AU validation set. }

\section{Introduction}
{As an important part of Artificial Intelligence and Human Interaction,  affective computing has arising more and more attention. Meanwhile, it has lots of applications in many fields, such as customer satisfaction survey, financial anti-fraud, and psychological analysis, etc. }

{The 3th ABAW Competition 2022 is a large- scale in the wild emotion database which is held by Dimitrios Kollias\cite{kollias2022abaw}\cite{kollias2021distribution}\cite{kollias2021analysing}\cite{kollias2021affect}, etc. It provides Aff-Wild2 which consists of three emotional database including categorical expression (such as happy, angry, sad) , valence arousal and 12 facial action units. Aff-Wild2 has 564 videos downloaded from YouTube. There are variety in ethnics, poses and ages, etc.\cite{zafeiriou2017aff}\cite{kollias2019deep}\cite{kollias2019expression}\cite{kollias2019face}\cite{kollias2020analysing}

{Different from seven basic categorical expressions and valence arousal, action units(AU) describe facial muscle movements developed by Paul Ekman in 1970s\cite{ekman1978facial}. Action units usually have concurrence. For example, AU 25 (lips part ) an AU 26 (jaw drop) often occur at the same time.}

{In this paper, we address AU task in ABAW 2022. In section 3, we present our methods of data balancing, model structure and loss function design. In section 4, we give the details about the datasets we used, experiment settings and our model ensembling strategy.}

\section{Method}
\subsection{Data balancing}

{Facial action unit recognition is a multi-label visual task in deep learning. There usually exist label imbalance problem in multi-label task and data balancing is very difficult because of label concurrence. Several papers are proposed to solve this problem. Wu Tong designed loss functions\cite{wu2020distribution} to solve this problem. Other scientists use data sampling to alleviate data unbalance.} 

{In this paper, we make Aff-Wild2 AU dataset more balanced with ML-ROS method\cite{charte2015addressing}.Another method we try to alleviate data unbalance problem is  batch sampling. When we train a batch from the training set. We wish all 12 Action Units (negative and positive ) are included in a batch. First, we organize the training data with 12 Action Units labels. Then when training is on, we sample each batch from the organized data. By doing this we can get a more balanced dataset.}

\subsection{Model Structure}
We use IResnet100\cite{duta2021improved} as the backbone. In order to better extract facial feature information, We choose pretrained weights on the Glint360K dataset\cite{an2020partical_fc}.During the experiment, We found that increasing the texture feature information of the face is helpful for the classification of AU. Due to the increase of the network depth, the semantic features are more abundant but the texture features will be lost, so we added the FPN and SSH modules \cite{Deng_2020_CVPR}to increase the texture information and receptive field of the face.see Figure 1.

At the same time, we flatten the features of each layer passing through FPN\cite{lin2017feature} and SSH\cite{najibi2017ssh} modules, and splicing the features of each layer to output 512 dimensions of features through a fully connected layer.And in order to make the network pay more attention to a certain part, we added the Coordinate Attention module\cite{Hou_2021_CVPR} to the shallow and deep layers of the network.The experimental results show that the feature information obtained in this way contains more texture features, and the classification effect is better than the previous AU classification.

\begin{figure}[h]
    \centering
    \includegraphics[width=1\textwidth]{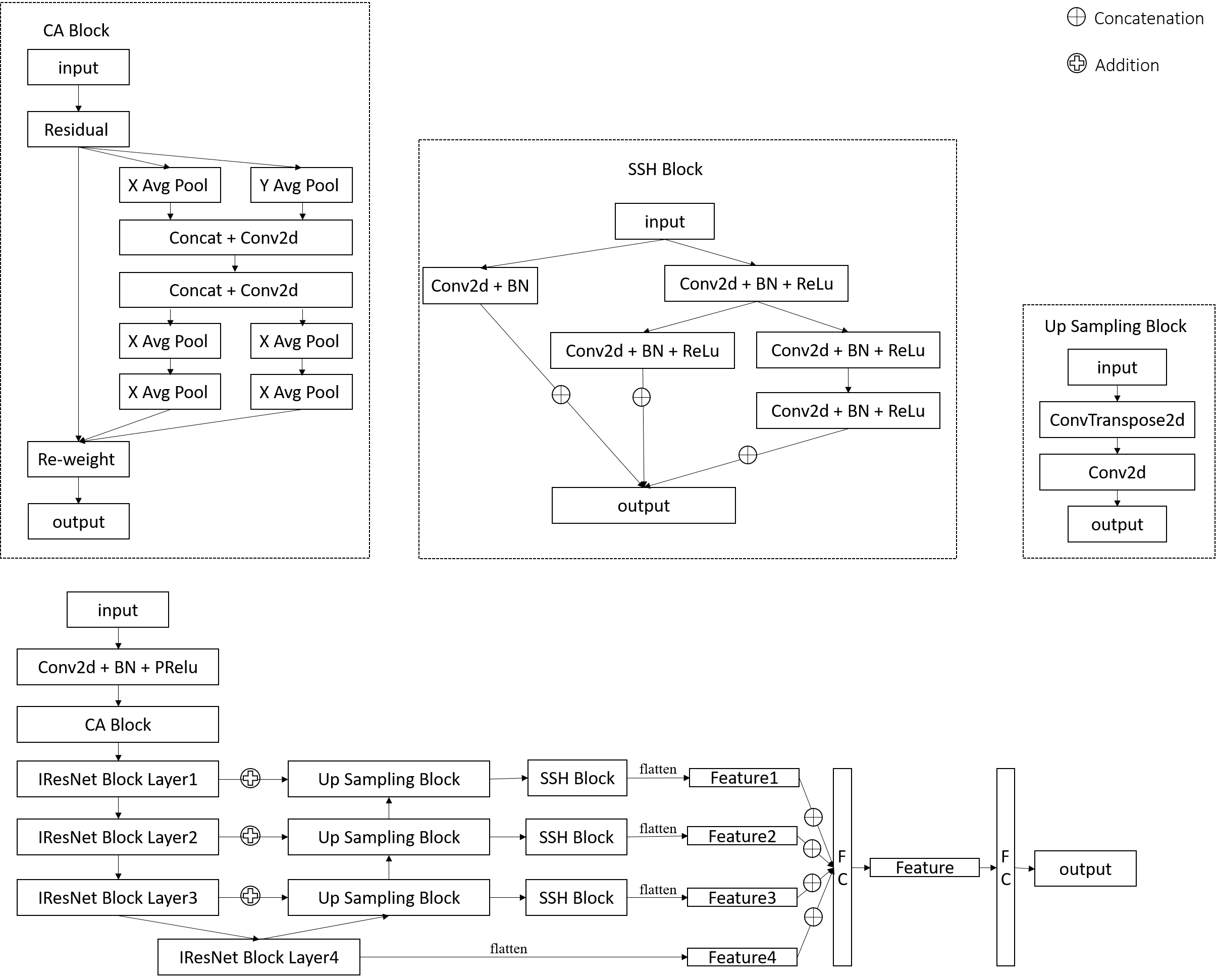}
    \caption{Overview system of proposed method}
    \label{fig:mesh2}
\end{figure}

\subsection{Loss Function}

For AU dataset in Aff-Wild2, we have counted the numbers of each AU in the training set and verification set, and the distribution of data set is shown in Figure 2. 

\begin{figure}[h]
    \centering
    \includegraphics[width=1\textwidth]{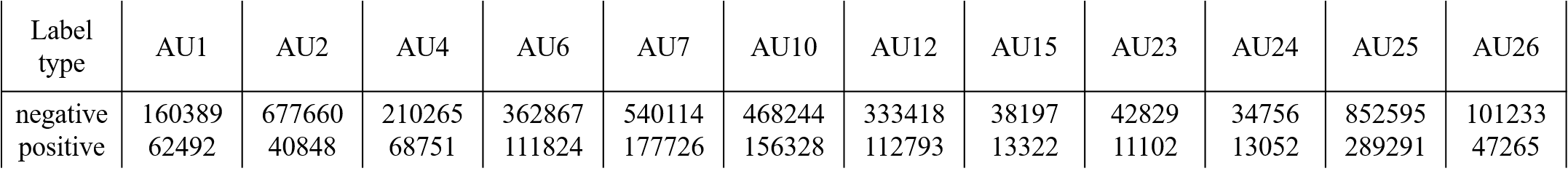}
    \caption{Numbers of positive and negative AU in both training set and validation set.}
    \label{fig:mesh2}
\end{figure}


As shown in Table 1, the data distribution of training set and verification set is extremely imbalanced, this will make AU15,AU23 and AU24 difficult to train. Because Action Unit Detection is a multi-label problem, Data Augment cannot solve the data imbalance problem. Therefore, we try to solve this problem from loss function.

\begin{equation}
bce\_loss(x,y) = \mathrm{L} = \{\mathrm{l}_1,...,\mathrm{l}_N\}.  \label{1} 
\end{equation}
where L represents the sum of the 12 AU, N represents the number of AU.

\begin{equation}
l_i = -\mathrm{w_n}[y_i \cdot \log\sigma(x_i) + (1 - y_i)\cdot \log(1 - \sigma(x_i))  \label{2} 
\end{equation}

\begin{equation}
W = \{\mathrm{w}_1,...,\mathrm{w}_N\} \label{3} 
\end{equation}

\begin{equation}
\sigma(x_i) = \frac{1}{1+\mathrm{e}^{-x_i})}^T  \label{4} 
\end{equation}

(2) is binary classification loss function, where W represents the loss weight of each AU, in our method, W = [1, 2, 1, 1, 1, 1, 1, 6, 6, 5, 1, 5], where x represents softmax output and in [0, 1] as (1), where y represents the target and takes either 0 or 1.

\begin{equation}
multi\_label\_loss(x,y) = -\mathrm{w_n} * \sum_{i}y[i] * \log((1 + \exp(-x[i]))^{-1}) + (1 - y[i])  \log\left(\frac{\exp(-x[i])}{(1+\exp(-x[i]))}\right) \label{5} 
\end{equation}

(5) is multi\_label loss, where x represents softmax output and in [0, 1], where y represents the target and takes either 0 or 1.

\begin{equation}
total\_loss = multi\_label\_loss(x,y) + ce\_loss(x,y) \label{6} 
\end{equation}

Finally, add bce\_loss and multi\_label\_loss together, as (6).

\subsection{Post Processing}
{Considering this au challenge needs video sequence predictions, we smooth the logits  generated by  the last layer of the network with a sliding window.}

\section{Experiments}

\subsection{Dataset}
Three datasets are used to train the backbone network as pretrained models respectively:

\textbf{Glint360K}: which is the largest and cleanest face recognition dataset and contains 170M images of 360k IDs, baseline models trained on Glint360K can easily achieve excellent performance.

\textbf{Private commercial EXPR dataset}: which contains 7K high definition images, each of which is human annotated into one of 7 facial expression categories(Neutral, Happy, Sad, Surprise, Fear, Disgust, Anger and Other).

\textbf{Private commercial AU dataset}: which contains 7K high definition images, each of which is human annotated into 15 face action unit categories(AU1, AU2, AU4, AU5, AU6, AU7, AU9, AU10, AU11, AU12, AU15, AU17, AU20, AU24 and AU26).

After we get the pretrained models trained from the above three datasets, we train AU dataset in Aff-Wild2\cite{kollias2019expression} to get our final results.
\subsection{Training and testing} In experiments, we use Iresnet100 to implement our framework. Our framework input size is 112x112. The SGD optimizer is used with a learning rate of 0.001, momentum of 0.9 and weight decay of 5e-4 and with a batch size of 256. The total training epoch is set as 15 in the ABAW training dataset. The learning rate will be divided by a factor of 10 when training to the 4/6/8 epoch. The data augment we implement Color Jitter(30\% chance of brightness, 30\% chance of contrast, 30\% chance of saturation and 30\% chance of hue), Random Horizontal Flip. Dropout of 0.6 is applied. All framework have been implemented in PyTorch, and training on 4 RTX-3090 GPU.

\begin{table}[h!]
  \begin{center}
    \begin{tabular}{c|c|c|c|c|c|c|c|c|c|c|c|c|c|c} 
     \textbf{r100} & \textbf{glint} & \textbf{bal} & \textbf{mll} & \textbf{ca} & \textbf{sa} & \textbf{ls} & \textbf{b+m} & \textbf{fpn} & \textbf{bs256} & \textbf{ssh} & \textbf{data} & \textbf{f1}\\
      \hline
      \checkmark &  &  &  &  &  &  &  &  &  &  &  & 0.390\\
      \checkmark & \checkmark &  &  &  &  &  &  &  &  &  &  & 0.534\\
      \checkmark & \checkmark & \checkmark &  &  &  &  &  &  &  &  &  & 0.549\\
      \checkmark & \checkmark & \checkmark & \checkmark &  &  &  &  &  &  &  &  & 0.570\\
      \checkmark & \checkmark & \checkmark & \checkmark & \checkmark &  &  &  &  &  &  &  & 0.614\\
      \checkmark & \checkmark & \checkmark & \checkmark & \checkmark & \checkmark &  &  &  &  &  &  & 0.662\\
      \checkmark & \checkmark & \checkmark & \checkmark & \checkmark & \checkmark & \checkmark &  &  &  &  &  & 0.673\\
      \checkmark & \checkmark & \checkmark & \checkmark & \checkmark & \checkmark & \checkmark & \checkmark &  &  &  &  & 0.690\\
      \checkmark & \checkmark & \checkmark & \checkmark & \checkmark & \checkmark & \checkmark & \checkmark & \checkmark &  &  &  & 0.709\\
      \checkmark & \checkmark & \checkmark & \checkmark & \checkmark & \checkmark & \checkmark & \checkmark & \checkmark & \checkmark &  &  & 0.712\\
      \checkmark & \checkmark & \checkmark & \checkmark & \checkmark & \checkmark & \checkmark & \checkmark & \checkmark & \checkmark & \checkmark &  & 0.715\\
      \checkmark & \checkmark & \checkmark & \checkmark & \checkmark & \checkmark & \checkmark & \checkmark & \checkmark & \checkmark & \checkmark & \checkmark & 0.721\\
      
    \end{tabular}
    \caption{All the effective methods are compared on the validation set.}
  \end{center}
\end{table}

\subsection{Ablation study}
We begin our oblation study by exploring the effectiveness of difference operation in our framework(Table 1). All experiments are based on Iresnet100(r100).Using glint360(glint) dataset  pretrain can improve by 14.4\%, using data balance(ba): oversampling AU2/AU15/AU23/AU24/AU26 and downsampling AU1/AU4/AU6/AU7/AU10/AU12/AU25 can improve by 1.5\%, using multi label loss(mll) can improve by 2.1\%, adding coordinate attention(ca) module can improve 4.4\%, using label smooth(ls) can improve 4.8\%, using bce loss + multi label loss(b+m) can improve 1.1\%, using feature pyramid networks(fpn) can improve 1.9\%, using bigger batch size 256 (bs256) can improve 0.3\%, using Single Stage Headless(shh) can improve 0.3\%, adding additional AU data can improve 0.6\%.

\subsection{Ensemble Model}
In previous competitions\cite{liu2018multi}\cite{liu2020group} we have used ensemble. In numerous experiments, we also adopted a model ensemble strategy, which ensemble the model with the highest F1 scale for each AU in the experiment, and obtained a higher F1 score on the validation set.see Figure 3.

\begin{figure}[h]
    \centering
    \includegraphics[width=1\textwidth]{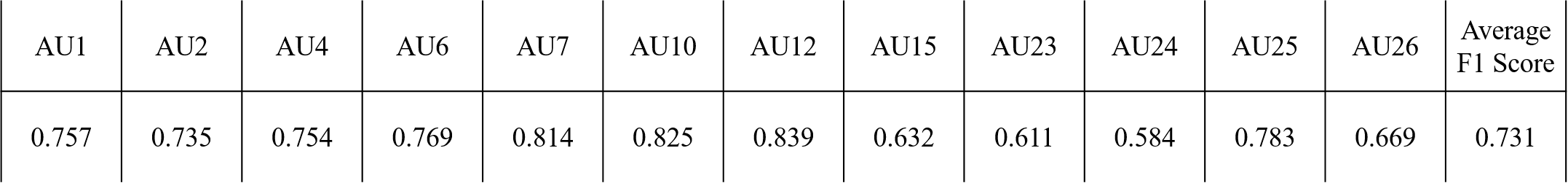}
    \caption{F1 score results for each AU}
    \label{figure:mesh2}
\end{figure}

\section{Conclusion}
For AU task in ABAW Competition 2022, we train the backbone on three different datasets and get three pretrained models respectively. Using multi label loss function, ML-ROS and batch sampling, the problem of data imbalancing is alleviated. Then we train Aff-Wild2 AU dataset on the three pretrained models. Finally, best checkpoint for each AU and each model are chosen and ensembled.
\printbibliography
\end{document}